\documentclass[10pt,twocolumn,letterpaper]{article}

\usepackage[pagenumbers]{cvpr} 

\usepackage{amsmath,amsfonts,bm}

\def\eqref#1{equation~\ref{#1}}

\def\1{\bm{1}}

\def\vtheta{{\bm{\theta}}}

\def\vphi{{\bm{\phi}}}

\def\vx{{\bm{x}}}
\def\vy{{\bm{y}}}

\DeclareMathAlphabet{\mathsfit}{\encodingdefault}{\sfdefault}{m}{sl}
\SetMathAlphabet{\mathsfit}{bold}{\encodingdefault}{\sfdefault}{bx}{n}

\def\gB{{\mathcal{B}}}
\def\gC{{\mathcal{C}}}

\def\gL{{\mathcal{L}}}

\def\gO{{\mathcal{O}}}

\def\gS{{\mathcal{S}}}

\def\gU{{\mathcal{U}}}

\def\gX{{\mathcal{X}}}
\def\gY{{\mathcal{Y}}}

\newcommand{\E}{\mathbb{E}}

\newcommand{\R}{\mathbb{R}}

\DeclareMathOperator*{\argmax}{arg\,max}

\usepackage{algorithm,algpseudocode}
\usepackage{amsthm}
\usepackage[table]{xcolor}
\usepackage{nicefrac}

\usepackage[most]{tcolorbox}
\definecolor{myboxcolor}{rgb}{0.402,0.402,0.402}
\newtcolorbox{contributions}[1][]{
        enhanced,
        title=#1,
        colback=myboxcolor!3,
        colbacktitle=myboxcolor!3,
        coltitle=black,
        left=4pt,
        right=4pt,
        top=4.5pt,
        bottom=0pt,
        attach boxed title to top center={xshift=0pt, yshift=-7pt},
        boxed title style={frame hidden, size=small, colback=myboxcolor!3},
        sharp corners,
        rounded corners,
        arc=7pt,
}

\def\methodname{{\upshape\textsc{Refine}}\xspace}
\newtheorem{theorem}{Theorem}

\newtheorem{definition}{Definition}

\usepackage{blindtext}

\definecolor{cvprblue}{rgb}{0.21,0.49,0.74}
\usepackage[pagebackref,breaklinks,colorlinks,allcolors=cvprblue]{hyperref}

\def\paperID{}
\def\confName{CVPR}
\def\confYear{2026}

\title{Cleaning the Pool: \\Progressive Filtering of Unlabeled Pools in Deep Active Learning}

\author{Denis Huseljic \quad Marek Herde \quad Lukas Rauch \quad Paul Hahn \quad Bernhard Sick  \\
      University of Kassel,  Germany \\
    {\tt\small \{firstname.lastname\}@uni-kassel.de}
}

\begin{document}
    \maketitle
    \begin{abstract}
Existing active learning (AL) strategies capture fundamentally different notions of data value, e.g., uncertainty or representativeness. Consequently, the effectiveness of strategies can vary substantially across datasets, models, and even AL cycles. Committing to a single strategy risks suboptimal performance, as no single strategy dominates throughout the entire AL process. We introduce \methodname, an ensemble AL method that combines multiple strategies without knowing in advance which will perform best. In each AL cycle, \methodname operates in two stages: (1) Progressive filtering iteratively refines the unlabeled pool by considering an ensemble of AL strategies, retaining promising candidates capturing different notions of value. (2) Coverage-based selection then chooses a final batch from this refined pool, ensuring all previously identified notions of value are accounted for. Extensive experiments across 6 classification datasets and 3 foundation models show that \methodname consistently outperforms individual strategies and existing ensemble methods.  Notably, progressive filtering serves as a powerful preprocessing step that improves the performance of any individual AL strategy applied to the refined pool, which we demonstrate on an audio spectrogram classification use case. Finally, the ensemble of \methodname can be easily extended with upcoming state-of-the-art AL strategies.
\end{abstract}    
    \section{Introduction}\label{sec:intro}
Pretrained foundation models have become the backbone of modern machine learning, offering powerful general-purpose representations~\cite{simeoni2025dinov3}. Yet, adapting these models to downstream tasks still often requires large amounts of annotated data. Obtaining annotations is costly and can quickly become a bottleneck in many applications. Active learning (AL) aims to mitigate this through an intelligent selection of instances for annotation~\cite{settles2009active}. Considering a pool-based batch AL setting, AL strategies query multiple instances per cycle from a large unlabeled pool, gradually expanding the training set while reducing annotation costs.

One central challenge in AL is identifying which instances maximize model performance. In recent years, much progress has been made in developing various selection strategies~\cite{ash2020deep,ash2021gone,bae2025uncertainty}. However, recent benchmarks show that their effectiveness varies substantially across models, domains, and even cycles of the AL process~\cite{munjal2022towards,rauch2023activeglae,luth2024navigating}. Moreover, choosing a suboptimal strategy can even yield worse performance than simple random sampling~\cite{munjal2022towards}. This variability poses a significant challenge for practitioners seeking to select an appropriate strategy for their downstream task, especially since AL is essentially a one-shot problem with little opportunity for trial and error. 

\begin{figure}
    \centering
    \begin{subfigure}[b]{\linewidth}
        \centering
        \includegraphics[width=.95\linewidth]{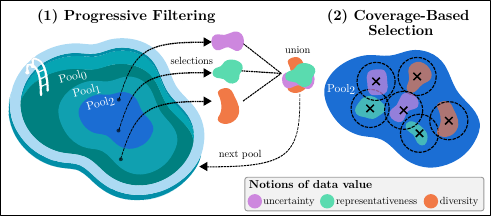}
        \caption{Conceptual illustration of our two-stage selection process.}
        \label{fig:sub_a}
    \end{subfigure}

    \vspace{2pt}
    
    \begin{subfigure}[b]{\linewidth}
    \centering
    \includegraphics[width=.95\linewidth]{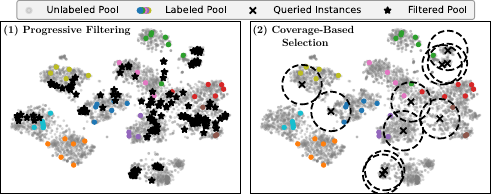} 
        \caption{t-SNE visualization on CIFAR-10 using DINOv2.}
        \label{fig:sub_b}
    \end{subfigure}

    \caption{\methodname's two-stage selection process. Stage 1: Progressive filtering refines the unlabeled pool through multiple rounds, where each round outputs the union of instances selected by each strategy as the next round's input pool. Stage 2: Coverage-based selection then chooses the final batch from this refined pool, ensuring a selection that accounts for diverse notions of value.}
    \label{fig:method_overview}
\end{figure}
Part of the challenge is that different selection strategies capture fundamentally different notions of data value, \eg,~uncertainty-based strategies target difficult instances while representativeness-based strategies seek high-density ones. Each perspective offers reasonable insights, yet no single strategy dominates across the entire AL process, as instances' impact on model performance strongly varies per cycle~\cite{hacohen2023select,bae2025uncertainty}. As the model learns, the optimal selection criterion shifts:~a strategy that excels in one AL cycle may become suboptimal in the next. This means that committing to any single strategy requires accepting its blind spots. While finding a universally optimal strategy is challenging, we presume that combining multiple complementary strategies in an \emph{ensemble AL method} provides a more robust alternative~\cite{huseljic2025boss}. While leveraging collective strengths, we reduce the risk of committing to any suboptimal heuristic.

\textbf{Our approach.}~We propose \methodname, an ensemble AL method that avoids committing to a single strategy. \methodname builds on the core insight that combining multiple complementary strategies, each considering its own notion of data value, can yield an improved selection that captures diverse perspectives of what makes an instance valuable.

Specifically, in each AL cycle, \methodname operates in two stages (\cref{fig:method_overview}):
\textbf{(1)~Progressive filtering} iteratively refines the unlabeled pool through multiple rounds. In each round, we apply all strategies from the ensemble on the current unlabeled pool and take the union of the selected instances, retaining batches that at least one strategy considers valuable. Repeating this process yields a high-value candidate pool, \ie, it reflects complementary notions of data value while filtering out uninformative instances that provide no value according to any strategy. This pool cleaning process is a powerful preprocessing step on its own, already considerably improving the performance of individual strategies.
\textbf{(2) Coverage-based selection} then selects the final batch from the refined pool by choosing the instances that, when combined with the already labeled pool, maximize coverage over the refined pool. Since progressive filtering has already identified high-value candidates, this selection ensures the batch is accounting for the various notions of data value captured in the filtering stage.

\noindent
Our contributions can be summarized as follows:
\begin{itemize}
    \item We introduce \emph{progressive filtering} to iteratively refine the unlabeled pool by retaining valuable instances while cleaning out uninformative ones, thereby reducing pool size and improving its quality for downstream selection.
    \item We propose \methodname, an \emph{ensemble AL method} that combines progressive filtering with a final coverage-based selection to leverage complementary notions of data value captured by multiple AL strategies in the refined pool.
    \item We provide a \emph{theoretical analysis} of progressive filtering, establishing bounds on its ability to retain high-value instances and remove uninformative ones.
    \item We demonstrate that \methodname \emph{outperforms} individual strategies and existing ensemble AL methods across 6 image classification datasets and 3 foundation models. Additionally, we validate the practicality of progressive filtering on an audio spectrogram classification use case.
\end{itemize}
    \section{Related Work}\label{sec:related_work}
\textbf{AL strategies} can be categorized into at least one of two criteria: uncertainty or representativeness. \textit{Margin}~\cite{settles2009active} focuses on uncertainty by selecting the top-$k$ instances with the smallest difference between the two most probable classes. \textit{TypiClust}~\cite{hacohen2022active} prioritizes representativeness via high-density instances, employing $k$-\textsc{means} for batch diversity. Similarly, \textit{BADGE}~\cite{ash2020deep} targets uncertainty through large gradient updates, utilizing $k$-\textsc{means++} to diversify the batch. \textit{BAIT}~\cite{ash2021gone} greedily selects instances maximizing Fisher information. Both \textit{AlfaMix}~\cite{parvaneh2022active} and \textit{DropQuery}~\cite{gupte2024revisiting} use perturbations (mixed features or dropout) to filter instances based on prediction variability, followed by $k$-\textsc{means} clustering. Finally, \textit{MaxHerding}~\cite{bae2024generalized} greedily maximizes generalized coverage via feature-based kernel similarities, which \textit{UHerding}~\cite{bae2025uncertainty} extends by incorporating an uncertainty factor.

\textbf{Ensemble AL methods} use multiple AL strategies by either \emph{combining them} or \emph{choosing when to use which}. Although \textit{ALBL}~\cite{hsu2015active}, \textit{COMB}~\cite{baram2004online}, and \textit{DUAL}~\cite{donmez2007dual} fit this regime, we do not consider them as they are single instance strategies whose adaptation to the batch setting is not straightforward. \textit{SelectAL}~\cite{hacohen2023select} chooses between low- and high-budget strategies by evaluating which one identifies the most valuable batch of the current labeled pool. However, consistent with \citet{bae2025uncertainty}, we found it yields inconsistent performance and is difficult to reproduce. \textit{TAILOR}~\cite{zhang2023algorithm} frames the choice of AL strategies as a multi-armed bandit problem, learning a probability distribution used to sample which strategies propose instances. As TAILOR uses a class-balance-based reward to update this distribution, it can handle class-imbalanced settings effectively. \textit{TCM}~\cite{doucet2024bridging} initially applies TypiClust for representativeness before transitioning to Margin for uncertainty, determining the switch point via a budget-dependent heuristic. This design is based on empirical findings that representative and diverse instances perform well on low budgets, while uncertainty is well-suited for high budgets~\cite{hacohen2023select}. Finally, \textit{AutoAL}~\cite{wang2025autoal} learns to weight the votes of multiple strategies via a model and selects the top-$k$ instances based on the total votes.

Unlike previous ensemble AL methods that attempt to learn or plan to switch between strategies, we propose \emph{progressive filtering}, a robust and training-free method. Instead of selecting a single strategy, this mechanism iteratively refines the unlabeled pool to retain high-value candidates from all available AL strategies. Furthermore, unlike strategies such as DropQuery or AlfaMix that filter in one step based on a single fixed heuristic, we generalize this concept by filtering across multiple rounds using diverse notions of data value. We provide both a theoretical analysis and empirical validation for the effectiveness of this concept.
    \section{Notation}\label{sec:notation}
We consider a pool-based batch AL setting for classification. Let $\vx \in \gX$ be an instance and $y \in \gY = \{1, \dots, K\}$ denote its label, where $K$ is the number of classes. Further, let $\gU_t \subset \gX$ be a large unlabeled pool and $\gL_t \subset \gX \times \gY$ be the labeled pool at cycle $t$. Let $\gS = \{s_1, \dots, s_M\}$ denote the set of $M$ selection strategies in the ensemble. At $t=0$, we initialize $\gL_0$ by randomly sampling $b$ instances. Then, we perform a total of $A$ AL cycles, selecting $b$ instances to label in each cycle. The total labeling budget is $B = b + A\cdot b$. We consider foundation models (or backbones) consisting of feature extractor $h^\vphi: \gX \to \R^D$ and classification head $g^\vtheta : \R^D \to \R^K$, where $\vphi$ and $\vtheta$ are fixed and trainable parameters, respectively. Hence, our model is a function $f = (g^\vtheta \circ h^\vphi)(\vx)$ mapping an instance to the logit space.
    \section{Method}\label{sec:method}
In this section, we detail \methodname with its two-stage selection process. We first introduce progressive filtering to refine the unlabeled pool for an improved downstream selection. We then describe how to effectively sample from this refined pool, accounting for the added value of each strategy. Afterward, we analyze theoretical properties of progressive filtering, demonstrating that it preserves high-value candidates while filtering out uninformative ones.

\subsection{Progressive Filtering}
We define the \emph{value} of an instance as its potential to improve model performance. What constitutes value depends on a strategy's perspective, such as reducing uncertainty or ensuring representativeness. An instance is considered \emph{collectively valuable} if it is ranked highly by at least one strategy, representing value from at least one perspective.

In each cycle, progressive filtering iteratively cleans the unlabeled pool to identify a subset of candidates that are collectively valuable across multiple selection strategies.  Our core idea is to let selection strategies act as experts, each providing a distinct perspective on data value.  When an expert considers a batch to be valuable, we include it in the refined pool. Through multiple iterations, this process progressively prioritizes high-value instances while filtering out uninformative ones. Intuitively, while experts may not agree on what is valuable, instances never selected by any expert are likely to be uninformative.

Formally, given the unlabeled pool $\gU_t$ at cycle $t$ and the set of $M$ selection strategies $\gS$, our goal is to obtain a candidate pool $\gC_R \subset \gU_t$ through $R$ rounds. In each round $r \in \{1, \dots, R\}$, each of the $M$ strategies select $J$ batches from the current refined pool $\gC_{r-1}$. By selecting multiple batches ($J > 1$), we prevent the pool $\gC_r$ from shrinking too rapidly and give each strategy multiple chances to identify valuable instances. To ensure computational tractability and to enable deterministic strategies to produce non-identical batches, each strategy $s_m$ selects from a randomly sampled subset of $\gC_{r-1}$. The refined pool for round $r$ is then the union of all selected batches:
\begin{align}\label{eq:union}
    \gC_r = \bigcup_{m=1}^{M} \bigcup_{j=1}^{J} s_m(\text{SubSample}(\gC_{r-1}, \alpha\cdot|\gC_{r-1}|), b), 
\end{align}
where $\gC_0 = \gU_t$, $\alpha \in (0, 1)$ is the sample ratio, and $\text{SubSample}(\cdot)$ draws instances uniformly without replacement. 
After $R$ rounds, the candidate pool $\gC_R$ serves as the refined pool from which the final batch for annotation is selected. The approach is summarized in \Cref{alg:filtering}.
\begin{algorithm}[t]
\small
\caption{Progressive Filtering}
\label{alg:filtering}
\begin{algorithmic}[1]
\Require Unlabeled pool $\gU_t$, selection strategies $\gS = \{s_1, \dots, s_M\}$, number of rounds $R$, number of batches per strategy $J$, batch size $b$, sampling ratio $\alpha$
\State $\gC_0 \gets \gU_t$
\For{$r \in \{1,\dots, R\}$}
    \State $\gC_r \gets \emptyset$
    \For{$m \in \{1, \dots, M\}$}
        \For{$j \in \{1, \dots, J\}$}
            \State $\gU_{\text{sample}} \gets \text{SubSample}(\gC_{r-1}, \alpha \cdot |\gC_{r-1}|)$
            \State $\gC_r \gets \gC_r \cup s_m(\gU_{\text{sample}}, b)$
        \EndFor
    \EndFor
\EndFor
\State \Return $\gC_R$
\end{algorithmic}
\end{algorithm}

\textbf{Design Rationale.} Three key design choices underpin progressive filtering.
\emph{Union Over Intersection:} While intersection might seem appealing for finding consensus instances, it has two critical limitations: (i) it discards instances that are uniquely valuable to specific strategies, reducing diversity, and (ii) it can result in (near-)empty sets when strategies have little overlap. The union operation preserves all potentially valuable instances while filtering uninformative instances through repetition. 
\emph{Iterative Refinement:} A single filtering round would simply concatenate all strategies' selections, providing only a slight reduction in uninformative instances. 
The power of progressive filtering emerges through multiple rounds: instances must be repeatedly selected to remain in the pool. Since each round operates on the output of the previous round, uninformative instances are unlikely to be consistently ranked high. Survival of instances through rounds requires consistent recognition of value. 
\emph{Sample Ratio:} The sample ratio $\alpha$ controls the stochasticity of the filtering process. Setting $\alpha$ close to $1$ makes the selection process more deterministic, which can cause the pool to shrink too rapidly. Reducing $\alpha$, we introduce controlled randomness that: (i) enables deterministic strategies to produce diverse batches, (ii) amplifies diversity by forcing exploration across different regions of the candidate space, and (iii) creates robust filtering where valuable instances have multiple chances to be selected. Furthermore, a smaller value for $\alpha$ reduces the computational and memory cost of selection strategies by limiting the pool size. In our experiments, we found that almost any value of $\alpha$ works well, provided it is not too small (\cf \cref{sec:experiments}).

\textbf{Practical Considerations.} We now discuss key implementation choices for applying progressive filtering. \emph{Strategy Set Composition:} The composition of the set $\gS$ represents a key design choice. Rather than attempting to find an optimal subset of strategies, which might be dataset- or model-dependent, we adopt a simple and principled approach: employing all individual batch AL strategies available in our benchmark in \cref{sec:experiments}. Progressive filtering places no restrictions on the types of AL strategies in its ensemble, requiring only that each can produce a batch of candidate instances. By incorporating all available perspectives, our method naturally captures diverse notions of value without manual tuning and prior knowledge. Moreover, this also makes progressive filtering easily expandable, allowing the ensemble to be extended with novel AL strategies that may offer new perspectives on instances' values.
\emph{Minimum Pool Size:} We enforce a minimum pool size equal to the batch size ($|\gC_R| \ge b$) to ensure that the subsequent coverage-based selection has sufficient instances for final selection. Moreover, the parameter $\alpha$ also mitigates this risk by introducing randomness into selections, preventing excessive shrinkage of $\gC_R$.
\emph{Hyperparameters:} We set $\alpha=0.4$, $R = 5$, and $J = 10$ based on a grid search on a validation split of CIFAR-10 and fix these values across all experiments. We further analyze their impact in \cref{sec:experiments}.

\textbf{Computational and Memory Efficiency.} Progressive filtering incurs additional computational overhead compared to standard batch selection, but this cost is justified by improved selection quality. 
The computational complexity per round is $\gO(M \cdot J \cdot C_{\text{max}}(\alpha \cdot |\gC_{r-1}|))$, where $C_{\text{max}}(\gC)$ is the cost of applying the most computationally expensive strategy on pool $\gC$. Two factors help control costs: (i) the sampling ratio $\alpha < 1$ reduces the pool size each strategy must evaluate, and (ii) ${\gC}_r$ is guaranteed to be a subset of $\gC_{r-1}$, ensuring the pool size decreases monotonically. This design also addresses a common memory bottleneck. Many AL strategies do not scale to extensively large pools, often requiring a fixed subsample of $\gU_t$ to be applicable. In contrast, by repeatedly using subsets of size $\alpha \cdot |\gC_{r-1}|$, progressive filtering can explore the entire unlabeled pool without the large memory footprint of evaluating the entire unlabeled pool at once. 
Overall, while progressive filtering incurs overhead, it is a reasonable trade-off for the performance gains, especially since the execution of AL strategies can be parallelized in practice. Assuming $M$ concurrent processes, the cost of progressive filtering effectively reduces to $\mathcal{O}(J \cdot C_{\text{max}}(\alpha \cdot |\gC_{r-1}|))$ per round.

\subsection{Coverage-Based Selection}\label{sec:coverage}
Given the refined pool $\gC_R$ from progressive filtering, we now address how to select the final batch for annotation. At this stage, $\gC_R$ consists primarily of valuable instances identified through collective assessment by multiple strategies. The remaining challenge is to select a batch that effectively leverages these diverse perspectives.

We select instances that, when added to $\gL_t$, best represent the distribution of $\gC_R$. Specifically, we seek a batch $\gB^* \subset \gC_R$ that maximizes coverage~\cite{bae2024generalized, bae2025uncertainty} over $\gC_R$:
\begin{align}\label{eq:coverage}
    \gB^* = \argmax_{\gB \subset \gC_R,\, |\gB| = b}  \E_{\vx} \left[  \max_{\vx' \in (\gL_t \cup \gB)} k(\vx, \vx') \right],
\end{align}
where $k(\vx, \vx')$ is a kernel measuring the similarity between instances. Intuitively, maximizing \cref{eq:coverage} ensures that every instance in $\gC_R$ is well-represented by at least one instance in $\gB \cup \gL_t$. This way, we capture diverse notions of value through progressive filtering in the final selection. Considering foundation models, \cref{eq:coverage} is maximized by applying the kernel to features $h^\vphi(\vx)$ rather than to the raw instances.

There are several existing AL strategies that explicitly optimize for coverage, typically over the entire unlabeled pool~\cite{sener2018active,hacohen2022active,bae2024generalized,bae2025uncertainty}. In our implementation, we adopt UHerding~\cite{bae2025uncertainty}, which extends MaxHerding~\cite{bae2024generalized} by greedily maximizing coverage with an additional uncertainty factor that balances exploration and exploitation.  We found this factor to be beneficial as most of the strategies in $\gS$ already focus on representativeness. Consequently, the final pool $\gC_R$ has a stronger focus on representative instances. However, any coverage-based selection strategy, such as TypiClust~\cite{hacohen2022active}, could be employed.

\textbf{Design Rationale.} The core idea is that by first refining the pool through progressive filtering, coverage-based methods can focus on selecting diverse instances from a high-value candidate set rather than attempting to balance value and diversity over the entire unlabeled pool. Without this separation, a single strategy must typically simultaneously determine instance value and ensure batch diversity, which is a challenging multi-objective optimization problem. By operating on the refined pool $\gC_R$, we can assume most of the candidates are valuable and focus on maximizing the coverage. This is both more efficient, as $|\gC_R| \ll |\gU_t|$, and more effective, since progressive filtering has already eliminated most uninformative instances.

\textbf{Limitations.} The candidate set $\gC_R$ naturally reflects the collective emphasis of strategy set $\gS$. When $\gS$ contains more representativeness-based strategies, $\gC_R$ will have more representative instances. In such cases, incorporating uncertainty (as done in UHerding) helps to maintain an appropriate balance between exploration and exploitation. Conversely, when $\gS$ contains more uncertainty-based strategies, the coverage optimization itself may provide the necessary representativeness.

\subsection{Theoretical Analysis of Progressive Filtering}
We now analyze the key properties of progressive filtering. We show that, with high probability, it (i)~preserves value through union operations and (ii)~exponentially reduces uninformative instances via iterative filtering.  Furthermore, we show that it (iii) guarantees that the expected true value of the candidate pool does not decrease. Proofs and numerical examples can be found in Appendix~\ref{app:proofs}.

\textbf{Value Preservation.}
The union operation in \cref{eq:union} ensures that instances deemed valuable by at least one strategy are likely to be preserved in the candidate pool.
\begin{theorem}[Value Preservation Bound]\label{thm:value}
    Let $\vx \in \gC_{r-1}$ be an instance and let $p_{m,r}(\vx)$ be the probability that strategy $s_m \in \gS$ includes $\vx$ in a selected batch during round $r$, given that $\vx$ is present in the random subsample.

    The probability of $\vx$ surviving round $r$, $P_r(\vx) = \Pr(\vx \in \gC_r \mid \vx \in \gC_{r-1})$, is lower-bounded by the probability from the single strategy that is most likely to select $\vx$:
    $$P_r(\vx) \geq 1 - \left(1 - \alpha \cdot \max_{m \in \{1, \dots, M\}} p_{m,r}(\vx) \right)^J.$$
\end{theorem}
\noindent
\Cref{thm:value} quantifies how progressive filtering avoids accidentally discarding valuable instances. An instance $\vx$ valued by any strategy (high $\max_m p_{m,r}(\vx)$) has its survival probability lower-bounded by the $J$ independent trials of the strategy it values most. This ensures that instances valued by any strategy are highly likely to remain in the pool, even with random subsampling. Importantly, the hyperparameters $J$ and $\alpha$ offer a way to control this behavior. 

\textbf{Exponential Reduction of Uninformative Instances.} Progressive filtering leads to an exponential reduction of uninformative instances through successive refinement. We formalize the uninformativeness of an instance through a probabilistic model.
\begin{definition}[Uninformative Instance]\label{def:uninformative_instance}
    An instance $\vx \in \gU_t$ is $\epsilon$-uninformative if for all strategies $s_m \in \gS$ and rounds $r$, the probability that $\vx$ appears in a randomly selected batch, given $\vx$ is present in the random subsample, is at most $\epsilon$.
\end{definition}
\noindent
Intuitively, uninformative instances are those that rank consistently low across all strategies and rounds and are not selected unless by random chance.

\begin{theorem}[Exponential Reduction of Uninformative Instances]\label{thm:uninformative}
    Let $\vx$ be an $\epsilon$-uninformative instance. The probability that $\vx$ survives $R$ rounds of progressive filtering is bounded by: $$\Pr(\vx \in \gC_R) \leq \left(1 - (1 - \alpha\epsilon)^{MJ}\right)^R.$$
    For $\alpha\epsilon \ll \frac{1}{MJ}$ and large $R$, this probability decreases exponentially as $\gO((MJ\alpha\epsilon)^R)$.
\end{theorem}
\noindent
\Cref{thm:uninformative} quantifies how the iterative process exponentially filters out uninformative instances. Since an uninformative instance is by definition unlikely to be selected by any strategy, it must survive the filtering process repeatedly across all $R$ rounds to remain in the pool. Repeated filtering ensures that the survival probability for such an instance decreases exponentially with each round.

\textbf{Concentration Toward a High-Value Pool.} Beyond preserving valuable instances and filtering out uninformative ones, progressive filtering ensures that the value of the candidate pool does not decrease over rounds. We formalize this using the expected value of the instances in the pool.
\begin{theorem}[Concentration of Expected Value]\label{thm:exp_value}
Let $V(\vx)$ be the unknown true value of an instance $\vx$, and let $\E[V|\gC] = \frac{1}{|\gC|} \sum_{\vx \in \gC} V(\vx)$ be the expected average value of pool $\gC$.
Assume that all strategies $s_m \in \gS$ are more likely to select higher-value instances:
$$
V(\vx) > V(\vy) \implies p_{m,r}(\vx) \geq p_{m,r}(\vy) \quad \forall m, r
$$
Then, the expected average value of the candidate pool is monotonically non-decreasing with each round of filtering:
$$
\mathbb{E}[V \mid \gC_R] \geq \mathbb{E}[V \mid \gC_{R-1}] \geq \dots \geq \mathbb{E}[V \mid \gC_0],
$$
where $\gC_0 = \gU_t$.
\end{theorem}
\noindent
\Cref{thm:exp_value} guarantees that the pool's average value is monotonically non-decreasing, \ie, it is guaranteed to improve or stay the same with each round. The key assumption, that AL strategies favor higher-value instances, is natural for any reasonable selection heuristic, though it may not always hold in practice. This theorem provides the theoretical view for why progressive filtering concentrates the pool toward high-value instances. In each round, lower-value instances are filtered out at higher rates than higher-value ones, thereby shifting the distribution of the pool toward instances with greater true value. Together with the other theorems, this ensures efficient convergence to a high-quality candidate pool for the second stage.
    \section{Experimental Evaluation}\label{sec:experiments}
We evaluate \methodname on image classification tasks, focusing primarily on standard benchmarks (\eg, CIFAR-10). We also present a use case study on audio spectrogram classification in Appendix~\ref{app:audio}. We start by establishing the experimental setup, then demonstrate that \methodname achieves superior performance against state-of-the-art strategies, followed by ablation studies analyzing key design choices. Our implementation can be found at \url{https://github.com/dhuseljic/dal-toolbox}.

\subsection{Experimental Setup}\label{sec:setup}
\textbf{Datasets and AL Protocol.} We evaluate \methodname on 6 image classification datasets. Following the AL protocol of~\cite{huseljic2025efficient}, we run 20 cycles starting with a randomly sampled initial labeled pool of size $b$. The batch sizes per datasets were determined by ensuring that accuracy converges with random sampling. Our benchmark includes datasets with varying numbers of classes (from $10$ to $200$) to assess robustness across different complexity levels. As CIFAR-10 was used for determining $R$, $J$, and $\alpha$, the \emph{main results} report performance on the remaining datasets. Details are summarized in \cref{tab:datasets}.
\begin{table}[!ht]
    \scriptsize
    \centering
    \caption{Overview of datasets. CIFAR-10* was used for hyperparameter selection.}
    \label{tab:datasets}
    \setlength{\tabcolsep}{7pt}
    \begin{tabular}{lrrrr}
    \toprule
    \rowcolor{gray!30}
    \textbf{Dataset} & \textbf{\# Instances} & \textbf{\# Classes} ($K$) & \textbf{Batch Size ($b$)} \\
    \midrule
    \makebox[65pt][l]{CIFAR-10*~\cite{krizhevsky2009learning}}   & 50,000 & 10    & 10 \\
    \makebox[65pt][l]{Dopanim~\cite{herde2024dopanim}}          & 10,484 & 15    & 50 \\
    \makebox[65pt][l]{Snacks~\cite{snacks2023dataset}}          & 4,838 & 20    & 20 \\
    \makebox[65pt][l]{CIFAR-100~\cite{krizhevsky2009learning}}  & 50,000 & 100   & 100 \\
    \makebox[65pt][l]{Food101~\cite{bossard14}}                 & 75,750 & 101   & 100 \\
    \makebox[65pt][l]{Tiny ImageNet~\cite{le2015tiny}}          & 100,000 & 200   & 200 \\
    \bottomrule
    \end{tabular}
\end{table}

\textbf{Models.} We employ 3 vision foundation models, each appended with a randomly initialized fully-connected classification head. We use DINOv2-ViT-S/14~\cite{oquab2024dinov2}, DINOv3-ViT-S/16~\cite{simeoni2025dinov3}, and CLIP-ViT-B/16~\cite{radford2021learning}. After each batch selection, we fine-tune the classification head for $200$ epochs while keeping the foundation model frozen. We use SGD with a batch size $64$, learning rate $0.01$, weight decay $10^{-4}$, and cosine annealing scheduling. These values were chosen to ensure consistent convergence across all datasets.

\textbf{Baseline Strategies.} We compare \methodname against random sampling and 8 state-of-the-art AL strategies. Our baselines include: (i)~the \emph{uncertainty-based} strategies Margin~\cite{settles2009active} and BADGE~\cite{ash2020deep}, (ii)~the \emph{representativeness-based} strategies TypiClust~\cite{hacohen2022active} and MaxHerding~\cite{bae2024generalized}, and (iii)~the \emph{hybrid} approaches BAIT~\cite{ash2021gone,huseljic2024fast}, ALFAMix~\cite{parvaneh2022active}, DropQuery~\cite{gupte2024revisiting} and UHerding~\cite{bae2025uncertainty}.
Finally, we include SelectAL~\cite{hacohen2023select}, TCM~\cite{doucet2024bridging}, TAILOR~\cite{zhang2023algorithm}, and AutoAL~\cite{wang2025autoal}, recent ensemble AL methods that combine multiple AL strategies.

\textbf{Evaluation Metrics.} We assess AL performance using two metrics: (i)~\emph{relative learning curves}, showing accuracy gains (absolute percentage points) over random sampling (from $\gU_t$) at each cycle, and (ii)~\emph{area under the learning curve} (AULC), measuring cumulative performance across all cycles. Absolute learning curves are provided in Appendix~\ref{app:complete_lcs}. All results are averaged over $10$ independent trials with shaded areas representing standard errors.
Additionally, we report \emph{pairwise win rates} between strategies: the percentage of trials where strategy $i$ achieves higher AULC than strategy $j$ across all cycles.

\begin{figure}[!t]
    \centering
    \includegraphics[width=\linewidth]{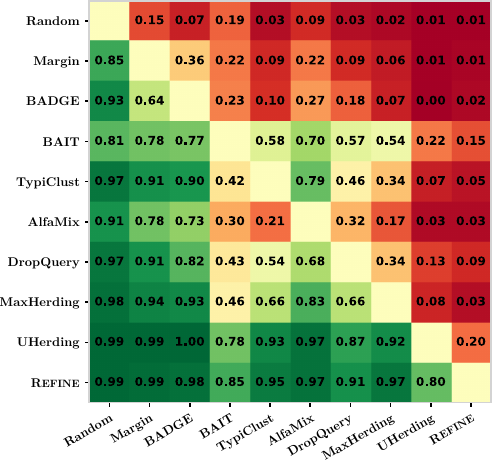}
    \caption{Pairwise comparison matrix averaged across 3 backbones, 5 datasets, and 10 trials. Element $(i,j)$ corresponds to the proportion of total runs, where strategy $i$ outperforms strategy $j$.}
    \label{fig:pairwise}
\end{figure}
\begin{figure*}[!t]
    \centering
    \includegraphics[width=.95\linewidth]{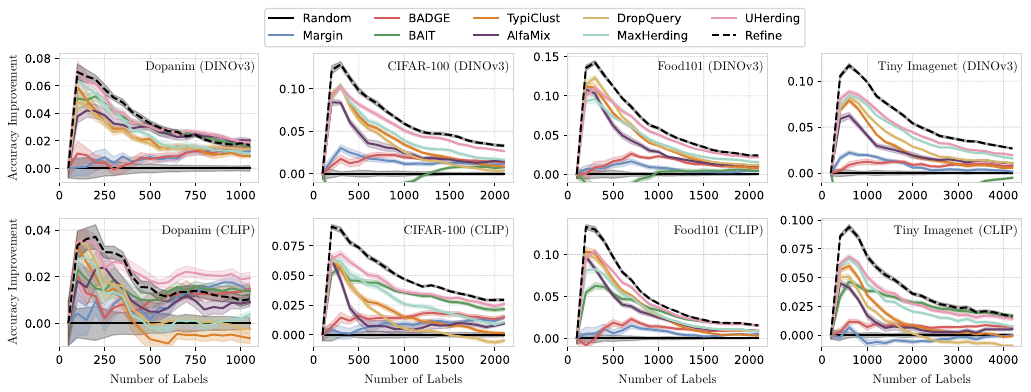}
    \caption{Relative accuracy learning curves for \methodname and baseline strategies across multiple backbones and datasets.}
    \label{fig:learning_curves}
\end{figure*}
\subsection{Main Results}\label{sec:main_results}
\textbf{Performance against Individual Strategies.}
\Cref{fig:pairwise} averages the results across all backbones and datasets in a pairwise comparison matrix, where each element $(i, j)$ reports the proportion of trials in which strategy $i$ outperforms strategy $j$. Overall, across all backbones and datasets, \methodname emerges as the most competitive strategy, achieving the highest win rates and outperforming every other strategy. 
For example, compared to other top-performing AL strategies like BAIT and UHerding, \methodname wins in 85\% and 80\% of all cases, respectively.

Additionally, \cref{fig:learning_curves} shows an exemplary subset of relative learning curves on the four most complex datasets, focusing on DINOv3 as the newest model and CLIP for its differing training process (\ie, student-teacher vs.~contrastive learning). The complete set of learning curves (relative and absolute) can be found in Appendix~\ref{app:complete_lcs}.  We observe that \methodname consistently outperforms all strategies on CIFAR-100, Food101, and Tiny ImageNet across all backbones, demonstrating its robustness across different datasets and models.

Importantly, we also want to highlight a limitation of progressive filtering: the performance can degrade when the majority of strategies in its ensemble perform poorly. This can be seen on the Dopanim dataset using the CLIP backbone, where progressive filtering does not yield the best learning curve.
Specifically, at around 350 labels, the accuracy of \methodname decreases slightly, which occurs as three AL strategies begin to perform equal to or worse than random sampling. While this is a general problem of ensemble methods, it could potentially be mitigated through weighting of ensemble components, which we identify as a promising direction for future research.

\begin{figure}[!hb]
    \centering
    \includegraphics[width=.6\linewidth]{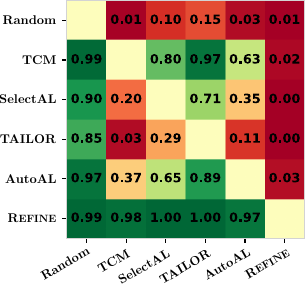}
    \caption{Pairwise comparison matrix of ensemble AL methods averaged across 3 backbones, 5 datasets, and 10 trials.}
    \label{fig:pairwise_ensemble}
\end{figure}

\textbf{Performance against Ensemble Methods.}
\Cref{fig:pairwise_ensemble} extends the pairwise comparison to ensemble AL methods. \methodname clearly outperforms all competitors, achieving a 100\% win rate against both SelectAL and TAILOR, 98\% against TCM, and 97\% against AutoAL, indicating that it combines AL strategies more effectively. Additional experiments and the complete set of associated learning curves can be found in Appendices~\ref{app:add_experiments} and~\ref{app:complete_lcs}, respectively.

\subsection{Ablation Studies}\label{sec:ablations}
We now investigate the contribution of \methodname's key components through four research questions, varying both datasets and backbones to ensure our findings are not specific to a particular experimental configuration. Additional results are provided in Appendix~\ref{app:ablations}.
\begin{table}[!ht]
\setlength{\tabcolsep}{2pt}
\centering
\scriptsize
    \caption{Effect of filtering rounds $R$, sampling ratio $\alpha$, and batches per strategy $J$ on pool quality. We report AULC improvement [\%] of random sampling when applied to filtered vs.~unfiltered pools.}
    \label{tab:ablation}
    \begin{subtable}[t]{0.32\linewidth}
        \centering
        \caption{Filtering rounds $R$}\label{tab:ablations_depth}
        \begin{tabular}{lrr}
        \toprule
        \rowcolor{gray!30}
        $R$ & CIFAR-10 & Snacks \\
        \midrule
        $1$ & 3.02 & 6.91 \\
        $2$ & 3.31 & 6.75 \\
        $3$ & 3.72 & 7.22 \\
        $5$ & 3.71 & 7.79 \\
        $7$ & 3.81 & 8.10 \\
        $9$ & 3.78 & 8.43 \\
        \bottomrule
        \end{tabular}
    \end{subtable}
    \begin{subtable}[t]{0.32\linewidth}
        \centering
        \caption{Sampling ratio $\alpha$}\label{tab:ablations_alpha}
        \begin{tabular}{lrr}
        \toprule
        \rowcolor{gray!30}
        $\alpha$ & CIFAR-10 & Snacks \\
        \midrule
        $0.1$ & 3.11 & 7.37 \\
        $0.2$ & 3.33 & 6.32 \\
        $0.3$ & 3.89 & 7.60 \\
        $0.5$ & 3.62 & 8.16 \\
        $0.7$ & 3.97 & 7.53 \\
        $0.9$ & 3.70 & 8.40 \\
        \bottomrule
        \end{tabular}
    \end{subtable}
    \begin{subtable}[t]{0.32\linewidth}
        \centering
        \caption{\# Batches $J$}\label{tab:ablations_J}
        \begin{tabular}{lrr}
        \toprule
        \rowcolor{gray!30}
        $J$ & CIFAR-10 & Snacks \\
        \midrule
        $1$  & 3.57 & 7.88 \\
        $5$  & 3.71 & 7.79 \\
        $10$ & 3.79 & 7.22 \\
        $12$ & 3.89 & 8.05 \\
        $15$ & 3.85 & 7.30 \\
        $20$ & 3.93 & 7.69 \\
        \bottomrule
        \end{tabular}
    \end{subtable}
\end{table}

\textbf{Does progressive filtering yield a higher-quality candidate pool?}
To isolate the contribution of progressive filtering from any specific AL strategy, we first apply filtering with $R \in \{1, 2, 3, 5, 7, 9\}$ and perform \emph{random sampling} from the filtered pool $\gC_R$. We fix $\alpha = 0.4$ and $J = 5$ throughout these experiments.
\Cref{tab:ablations_depth} shows that progressive filtering consistently improves pool quality across CIFAR-10 and Snacks. We observe better accuracies compared to random sampling from $\gU_t$, with performance increasing as more filtering rounds are applied. This can saturate with higher rounds, \eg, for CIFAR-10 and $R=9$, the improvement plateaus, indicating diminishing returns from additional filtering rounds. However, overall, increasing depth $R$ generally yields a higher-quality candidate pool.

\textbf{How do $\alpha$ and $J$ affect performance of progressive filtering?}
We again isolate the contribution of progressive filtering from any specific selection strategy and examine $\alpha$ and $J$.
\Cref{tab:ablations_alpha} shows that random sampling with progressive filtering ($R=5$ and $J=5$) consistently outperforms random sampling from $\gU_t$ across all values of $\alpha$. Values that are too low ($\alpha \le 0.2$) can slightly reduce filtering effectiveness, while higher values ($\alpha \in [0.3, 0.9]$) perform well for both datasets. Importantly, since $\alpha$ determines the pool size from which strategy $s_m$ samples, lower values are preferable as they reduce computational and memory cost. Similarly, \cref{tab:ablations_J} demonstrates that progressive filtering ($R=5$ and $\alpha=0.4$) generally improves performance when increasing $J$. Accuracy improves as more batches are selected per strategy, but this benefit can diminish for larger values.

\textbf{How beneficial is progressive filtering as a preprocessing step for AL strategies?}
\begin{figure}[!t]
    \centering
    \begin{subfigure}[b]{.98\linewidth}
        \centering
        \includegraphics[width=\linewidth]{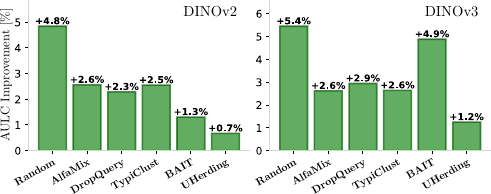}
        \caption{AULC improvement [\%] using DINOv2 and DINOv3.}
        \label{fig:aulc_improvments}
    \end{subfigure}
    \vspace{1em}
    
    \begin{subfigure}[b]{.98\linewidth}
        \centering
        \includegraphics[width=\linewidth]{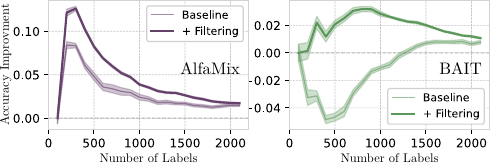}
        \caption{Relative curves for AlfaMix (+2.6\%) and BAIT (+4.9\%) on DINOv3.}
        \label{fig:relacc_strats}
    \end{subfigure}
    \caption{Benefits of progressive filtering (selecting from $\gC_R$ vs.~$\gU_t$) on CIFAR-100 across different strategies.}
    \label{fig:filtering_benefits}
\end{figure}
We now examine how progressive filtering benefits different AL selection strategies. In \methodname, we employ coverage-based selection on the candidate pool $\gC_R$ to efficiently cover valuable instances. Here, we isolate this second stage by first applying progressive filtering with fixed parameters ($R = 5$, $\alpha = 0.4$, $J = 5$), then comparing selection strategies from $\gS$ when applied to the filtered pool $\gC_R$ versus the unfiltered pool $\gU_t$.
\Cref{fig:aulc_improvments} shows AULC improvements for random sampling and five state-of-the-art selection strategies on CIFAR-100 using DINOv2 and DINOv3. Progressive filtering consistently improves all strategies for both backbones, with gains ranging from +0.7\% to +5.4\%, demonstrating that any selection strategy can benefit from our preprocessing step. Notably, weaker AL strategies benefit most from progressive filtering, while well-performing strategies like UHerding show smaller gains.
\Cref{fig:relacc_strats} shows exemplary learning curves for AlfaMix and BAIT on CIFAR-100 using DINOv3. While progressive filtering improves both strategies, the effect is critical for BAIT. Without filtering, BAIT (Baseline) performs even worse than random sampling. However, incorporating filtering improves its performance, showing the benefit of an ensemble of AL strategies. 

\begin{figure}[!t]
    \centering
    \includegraphics[width=\linewidth]{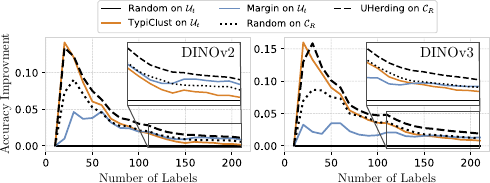}
    \caption{Relative learning curves on CIFAR-10 for strategies on $\gU_t$ and $\gC_R$ with filtering restricted to Margin and TypiClust.}
    \label{fig:ablation_value}
\end{figure}

\textbf{Does progressive filtering successfully combine complementary notions of value?}
To investigate whether progressive filtering properly captures different notions of value in the final pool $\gC_R$, we perform filtering with only two \emph{complementary} selection strategies: Margin, focusing on uncertain instances, and TypiClust, focusing on representative instances. 
If filtering captures these notions correctly, this should be evident when randomly sampling from $\gC_R$. We fix hyperparameters to the values used in our main experiments ($R=5$, $\alpha = 0.4$, $J = 10$). 
\Cref{fig:ablation_value} demonstrates that the final pool $\gC_R$ correctly captures both notions of value. Initially, TypiClust yields higher performances due to its focus on representative instances, and random sampling from $\gC_R$ effectively profits from this. As TypiClust's effectiveness diminishes in later cycles, random sampling from $\gC_R$ benefits from uncertain instances captured by Margin. This behavior is consistent across both backbones. Moreover, coverage-based selection via UHerding on $\gC_R$ further improves performance, demonstrating its ability to capture these diverse notions of value.

    \section{Conclusion}
We introduced \methodname, an ensemble AL method that combines different notions of value to achieve state-of-the-art performance. Our core contribution, progressive filtering, iteratively refines the unlabeled pool via the union of batches from multiple AL strategies. This process cleans the unlabeled pool by discarding uninformative instances, yielding a high-value candidate pool for downstream selection. A subsequent coverage-based selection effectively yields a diverse batch from this refined, high-quality pool.

\textbf{Limitations.} The effectiveness of \methodname depends on the quality and composition of its strategy ensemble. If these strategies are correlated (\eg, all focus on diversity), the filtered pool will inherit this bias. Moreover, the union can be sensitive if there are many poorly-designed strategies, which may slightly degrade the value of the candidate pool. We therefore recommend building $\gS$ from a small set of well-established, com\-ple\-men\-tary strategies rather than arbitrarily {including numerous ones.\par}
    {
        \small
        \bibliographystyle{ieeenat_fullname}
        \bibliography{main}
    }
    \appendix
    \clearpage
\setcounter{page}{1}
\maketitlesupplementary

\section{Proofs of Theorems}\label{app:proofs}
Here, we provide detailed proofs for the theorems from the main paper. Furthermore, to complement the theoretical analysis of \cref{thm:value,thm:uninformative}, we also include numerical examples illustrating the probability of preserving valuable instances and filtering out uninformative ones.

\textbf{Proof of \cref{thm:value}.} Let $P_r(\vx) = \Pr(\vx \in \gC_r \mid \vx \in \gC_{r-1})$ be the probability of instance $\vx$ surviving round $r$. Given $\vx \in \gC_{r-1}$, let $S_{m,r}$ be the \emph{event} that AL strategy $s_m$ selects $\vx$ in round $r$ at least once in its $J$ trials. The total survival event is the union of these individual events, \ie, $\vx$ survives if at least one strategy selects it:
$$P_r(\vx) = \Pr\left(\bigcup_{m=1}^M S_{m,r}\right).$$
We first derive $\Pr(S_{m,r})$ and then use this to find the lower bound for the union.
\begin{enumerate} 
    \item For a single draw $j\in\{1, \dots, J\}$ by strategy $s_m$, the event of selecting $\vx$ requires two \emph{independent conditions} to be met: 
    \begin{enumerate} 
        \item Instance $\vx$ must be present in the random subsample $\gU_{\text{sample}}$. The probability of this is 
        $$\Pr(\vx \in \gU_{\text{sample}}) = \alpha.$$
        \item Strategy $s_m$ must select $\vx$ from that subsample. By definition, this probability is $p_{m,r}(\vx)$. 
    \end{enumerate}
    \item The joint probability of $\vx$ being selected in a single draw $j$ is the product of these probabilities:
    $$\Pr(\text{select } \vx \text{ in draw } j) = \alpha \cdot p_{m,r}(\vx).$$
    \item The probability of $\vx$ \emph{not being selected} in this single draw is $1 - \alpha \cdot p_{m,r}(\vx)$.
    \item As per Algorithm~\ref{alg:filtering}, strategy $s_m$ makes $J$ independent draws (each with a new subsample). The probability that $\vx$ is \emph{never selected} by $s_m$ across all $J$ draws is the product of their individual probabilities:
    $$\Pr(\Bar{S}_{m,r}) = \prod_{j=1}^J (1 - \alpha \cdot p_{m,r}(\vx)) = (1 - \alpha \cdot p_{m,r}(\vx))^J.$$
    \item Therefore, the probability that $\vx$ is selected \emph{at least once} by strategy $s_m$ is the complement:
    $$\Pr(S_{m,r}) = 1 - \Pr(\Bar{S}_{m,r}) = 1 - (1 - \alpha \cdot p_{m,r}(\vx))^J.$$
    \end{enumerate}

    \noindent
    Having found $\Pr(S_{m,r})$, we now derive the lower bound.
    \begin{enumerate}
    \item The probability of a union of events is always greater than or equal to the probability of any single event in that union. Therefore, for any strategy $s_i$:
    $$
    P_r(\vx) = \Pr\left(\bigcup_{m=1}^M S_{m,r}\right) \ge \Pr(S_{i,r}).
    $$

    \item Since this holds for all $i$, it must also hold for the maximum probability among them:
    $$
    P_r(\vx) \ge \max_{m \in \{1, \dots, M\}} \Pr(S_{m,r}).
    $$

    \item Substituting the expression for $\Pr(S_{m,r})$ from Step 5:
    $$
    P_r(\vx) \ge \max_{m \in \{1, \dots, M\}} \left[ 1 - (1 - \alpha \cdot p_{m,r}(\vx))^J \right].
    $$

    \item The function $f(p) = 1 - (1 - \alpha p)^J$ is monotonically increasing with $p$. Therefore, the maximum value of the function is achieved when $p$ is at its maximum.
    
    Let $p_{\max} = \max_{m \in \{1, \dots, M\}} p_{m,r}(\vx)$. Then:
    $$
    \max_{m} \left[ 1 - (1 - \alpha \cdot p_{m,r}(\vx))^J \right] = 1 - (1 - \alpha \cdot p_{\max})^J.
    $$
    
    \item This proves the final bound:
    $$
    P_r(\vx) \ge 1 - \left(1 - \alpha \cdot \max_{m \in \{1, \dots, M\}} p_{m,r}(\vx) \right)^J. \quad \blacksquare
    $$
\end{enumerate}
\textbf{Numerical Example.} We provide a numerical example to illustrate the probability of preserving a valuable instance. We consider $\alpha = 0.4$, $J = 10$, and an instance $\vx$ that receives a high score from Margin ($p_\text{margin} = 0.9$) but a low score from TypiClust ($p_\text{typiclust} = 0.2$). We assume an ensemble of two strategies, where the instance's true value is correlated with the uncertainty heuristic, which is accounted for by the Margin. Since the bound uses the maximum score, we have $p_\text{max} = 0.9$. This yields $P_r(\vx) \geq 1 - (1 - 0.4 \cdot 0.9)^{10} = 1 - (0.64)^{10} \approx 0.9885$, demonstrating that the instance has a high lower-bound probability because at least one strategy strongly prioritizes it.

\textbf{Proof of \cref{thm:uninformative}.}
Let $\vx$ be an $\epsilon$-uninformative instance and let $P_r(\vx) = \Pr(\vx \in \gC_r \mid \vx \in \gC_{r-1})$ be the probability that an $\epsilon$-uninformative instance $\vx$ survives round $r$. We first derive an upper bound for $P_r(\vx)$ and then use this to find the bound for surviving all $R$ rounds.
\begin{enumerate} 
    \item For a single draw $j\in\{1, \dots, J\}$ by strategy $s_m$, the event of selecting $\vx$ requires \emph{two conditions} to be met: 
    \begin{enumerate} 
        \item Instance $\vx$ must be present in the random subsample $\gU_{\text{sample}}$. The probability of this is $$\Pr(\vx \in \gU_{\text{sample}}) = \alpha.$$
        \item Strategy $s_m$ must select $\vx$ from that subsample. By Definition \ref{def:uninformative_instance}, this probability is bounded: $p_{m,r}(\vx) \le \epsilon$. 
    \end{enumerate} 
    \item The joint probability of $\vx$ being selected in a single draw $j$ by strategy $s_m$ is bounded: 
    $$\Pr(\text{select } \vx \text{ in draw } (m,j)) = \alpha \cdot p_{m,r}(\vx) \le \alpha \epsilon.$$
    \item The probability of $\vx$ \emph{not} being selected in this single draw is therefore lower-bounded:
    \begin{align*}
    \Pr(\text{not select } \vx \text{ in draw } (m,j)) 
    &\ge 1 - \alpha \epsilon.
    \end{align*}
    \item As per Algorithm \ref{alg:filtering}, we make a total of $M \cdot J$ independent draws in round $r$. The event that $\vx$ is \emph{never selected} requires it to be missed by all $M \cdot J$ draws. Using the lower bound from the previous step, we get a lower bound for $\vx$ not surviving the round:
    $$
    \Pr(\vx \notin \gC_r \mid \vx \in \gC_{r-1}) \ge \prod_{m=1}^M \prod_{j=1}^J (1 - \alpha \epsilon) = (1 - \alpha \epsilon)^{MJ}.
    $$
    \item Consequently, the event that $\vx$ is selected at least once is bounded by:
    \begin{align*}
    P_r(\vx) &= \Pr(\vx \in \gC_r \mid \vx \in \gC_{r-1}) \\
    &= 1 - \Pr(\vx \notin \gC_r \mid \vx \in \gC_{r-1}) \\
    &\le 1 - (1 - \alpha \epsilon)^{MJ}.
    \end{align*}
\end{enumerate}

\noindent
Having found the upper bound for $P_r(\vx)$, we now derive the bound for surviving all $R$ rounds.
\begin{enumerate}
    \item For $\vx$ to survive all $R$ rounds, it must survive each round. The total survival probability is the product of the per-round survival probabilities:
    $$
    \Pr(\vx \in \gC_R) = \prod_{r=1}^R \Pr(\vx \in \gC_r \mid \vx \in \gC_{r-1}) = \prod_{r=1}^R P_r(\vx).
    $$
    \item Since, by definition, instances $\vx$ remain $\epsilon$-uninformative across rounds (for all $r$):
    \begin{align*}
        \Pr(\vx \in \gC_R) &\le \prod_{r=1}^R \left( 1 - (1 - \alpha \epsilon)^{MJ} \right)\\
        &= \left( 1 - (1 - \alpha \epsilon)^{MJ} \right)^R, 
    \end{align*}
    proving the first part of \cref{thm:uninformative}.
    
    \item For the approximation, we assume $\alpha\epsilon \ll \frac{1}{MJ}$. Using the approximation $(1-x)^n \approx 1 - nx$ for small $x$:
    $$
    (1 - \alpha \epsilon)^{MJ} \approx 1 - MJ\alpha\epsilon.
    $$
    \item Substituting this approximation back into the bound, we obtain that
    $$
    \Pr(\vx \in \gC_R) \lesssim \left( 1 - (1 - MJ\alpha\epsilon) \right)^R = (MJ\alpha\epsilon)^R,
    $$
    showing that the probability decreases exponentially with $R$. $\quad \blacksquare$
\end{enumerate}
\textbf{Numerical Example.} We provide a numerical example to illustrate the survival probability of an uninformative instance. We set $R = 5$, $M = 3$, $J = 10$, and $\alpha = 0.4$. Furthermore, we consider an $\epsilon$-uninformative instance $\vx$ with $\epsilon = 0.05$.  The probability that this instance survives a single round is 
\begin{align*}
  P_r(\vx) 
  &\le 1 - (1 - \alpha\epsilon)^{MJ} \\
  &= 1 - (1 - 0.4 \cdot 0.05)^{30} \\
  &= 1 - (0.98)^{30} \approx 0.4545.
\end{align*} 
Extending this over all $R=5$ rounds, the bound becomes $\Pr(\vx \in \gC_5) \leq (1 - (0.98)^{30})^5 \approx 0.0194$. After 5 rounds, the probability of this uninformative instance remaining in the candidate pool is bounded below $2\%$.

\textbf{Proof of \cref{thm:exp_value}.}
Let $V(\vx)$ be the unknown true value of instance $\vx$, and let $\E[V|\gC] = \frac{1}{|\gC|} \sum_{\vx \in \gC} V(\vx)$ be the expected average value of pool $\gC$.
We show that the expected value is non-decreasing in any single round $r$.
\begin{enumerate} 
    \item Let $\E[V \mid \gC_{r-1}] = \frac{1}{|\gC_{r-1}|} \sum_{\vx \in \gC_{r-1}} V(\vx)$ be the average value of the pool at the start of the round. 
    \item The pool $\gC_r$ is formed by sampling from $\gC_{r-1}$. As shown in the proof of \cref{thm:value}, the probability that an instance $\vx \in \gC_{r-1}$ survives to $\gC_r$ is 
    $$P_r(\vx) = \Pr(\vx \in \gC_r \mid \vx \in \gC_{r-1}).$$
    \item The expected value of the new pool $\gC_r$ is the weighted average of the values from the previous pool, where the weights are these survival probabilities\footnote{Since we iterate over the specific instances fixed in the previous pool $\gC_{r-1}$, the value $V(\vx)$ is treated as a constant.}:
    $$\mathbb{E}[V \mid \mathcal{C}_r] = \frac{\sum_{\vx \in \mathcal{C}_{r-1}} V(\vx) P_r(\vx)}{\sum_{\vx \in \mathcal{C}_{r-1}} P_r(\vx)}.$$
    \item From the theorem's assumption, an instance's value $V(\vx)$ is positively correlated with its selection probability $p_{m,r}(\vx)$ for all strategies:
    $$ V(\vx) > V(\vy) \implies p_{m,r}(\vx) \geq p_{m,r}(\vy) \quad \forall m, r .$$ 
    \item From this and the fact that $P_r(\vx)$ is a monotonically increasing function of each $p_{m,r}(\vx)$, it follows that $V(\vx)$ is \emph{also positively correlated} with $P_r(\vx)$:
    $$V(\vx) > V(\vy) \implies P_r(\vx) \ge P_r(\vy).$$
    \item Because the values $V(\vx)$ and the weights $P_r(\vx)$ are positively correlated, \emph{Chebyshev's sum inequality} shows that the weighted average must be greater than or equal to the unweighted average:
    $$ \underbrace{\frac{\sum_{\vx \in \mathcal{C}_{r-1}} V(\vx) P_r(\vx)}{\sum_{\vx \in \mathcal{C}_{r-1}} P_r(\vx)}}_{\mathbb{E}[V \mid \mathcal{C}_r]} \ge \underbrace{\frac{\sum_{\vx \in \mathcal{C}_{r-1}} V(\vx)}{|\mathcal{C}_{r-1}|}}_{\mathbb{E}[V \mid \mathcal{C}_{r-1}]} $$
    
    \item Since $\mathbb{E}[V \mid \mathcal{C}_r] \ge \mathbb{E}[V \mid \mathcal{C}_{r-1}]$ holds for any round $r$, applying this result iteratively proves:
    $$ \mathbb{E}[V \mid \mathcal{C}_R] \geq \mathbb{E}[V \mid \mathcal{C}_{R-1}] \geq \dots \geq \mathbb{E}[V \mid \mathcal{C}_0]. \quad \blacksquare$$
\end{enumerate}

\section{Use Case: Audio Spectrogram Classification}\label{app:audio}
This use case demonstrates the practical utility of progressive filtering as a preprocessing step in AL. Specifically, we demonstrate that practitioners can achieve substantial accuracy gains, even transforming a previously underperforming strategy into a viable one. They can accomplish this without discarding their existing, trusted AL workflows by employing progressive filtering as a workflow-agnostic method to refine the unlabeled pool. This approach enhances any AL workflow with minimal engineering effort and risk. An additional benefit can also be computational feasibility. By reducing the unlabeled pool size, progressive filtering enables the practical application of complex or computationally expensive AL strategies that would otherwise be prohibitively costly on large (unfiltered) unlabeled pools.

\begin{figure}[!ht]
    \centering
    \includegraphics[width=\linewidth]{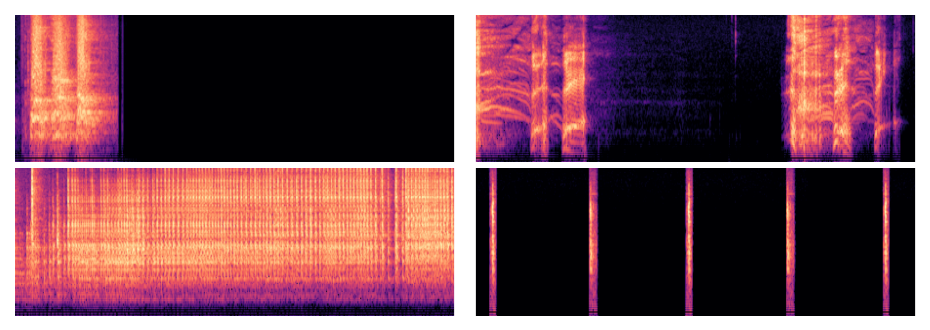}
    \caption{Example spectrograms from ESC50.}
    \label{fig:spectograms}
\end{figure}
To illustrate this, we conducted a use case study on the ESC50 audio dataset~\cite{piczak2015esc}. As ESC50 does not have a dedicated test split, we randomly sampled 20\% of the data for evaluation. We transformed audio segments into spectrograms following~\cite{rauch2025unmutepatchtokensrethinking} (\cf \cref{fig:spectograms}) and employed the EAT base backbone to extract features~\cite{chen2024_EAT}. Progressive filtering was applied exactly as described in the main paper, using the same hyperparameters and ensemble $\gS$. This use case simulates a practitioner whose trusted AL strategy is Coreset~\cite{sener2018active}, which is highly cited and well-known. Critically, Coreset is not part of the ensemble $\gS$ used for filtering, which is a scenario where progressive filtering enhances an external strategy.
\begin{figure}[!ht]
    \centering
    \includegraphics[width=\linewidth]{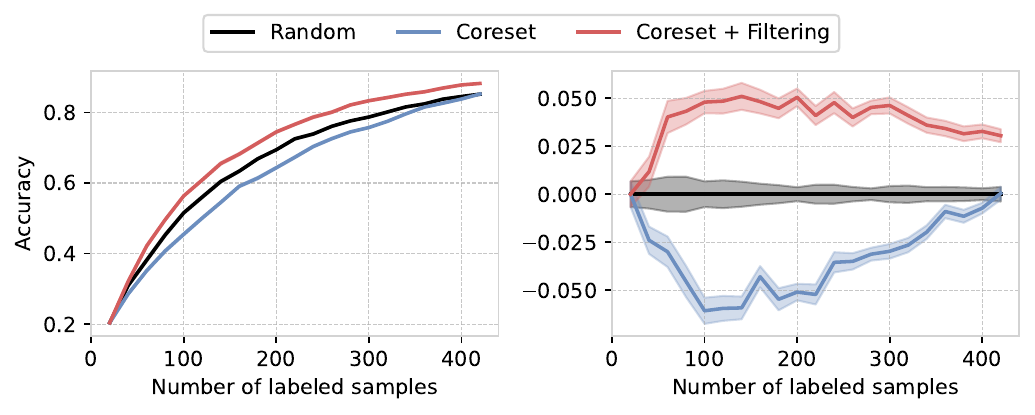}
    \caption{Absolute and relative learning curves comparing random sampling on $\gU_t$, Coreset on $\gU_t$, Coreset on $\gC_R$ on ESC50.}
    \label{fig:audio_lcs}
\end{figure}

\noindent
As shown in \Cref{fig:audio_lcs}, applying the original Coreset strategy to the entire unlabeled pool $\gU_t$ yields accuracy considerably below the random sampling baseline. In this instance, the standard strategy failed, meaning the practitioner would have obtained better results using simple random sampling. In contrast, by applying progressive filtering as a preprocessing step, we observe an accuracy improvement of up to 10\%. This demonstrates that filtering can even effectively ``fix'' a failing strategy, allowing it to outperform baselines (\eg, random) without requiring practitioners to abandon their preferred workflow.

\section{Additional Ablations}\label{app:ablations}
\begin{figure*}
    \centering
    \includegraphics[width=.95\linewidth]{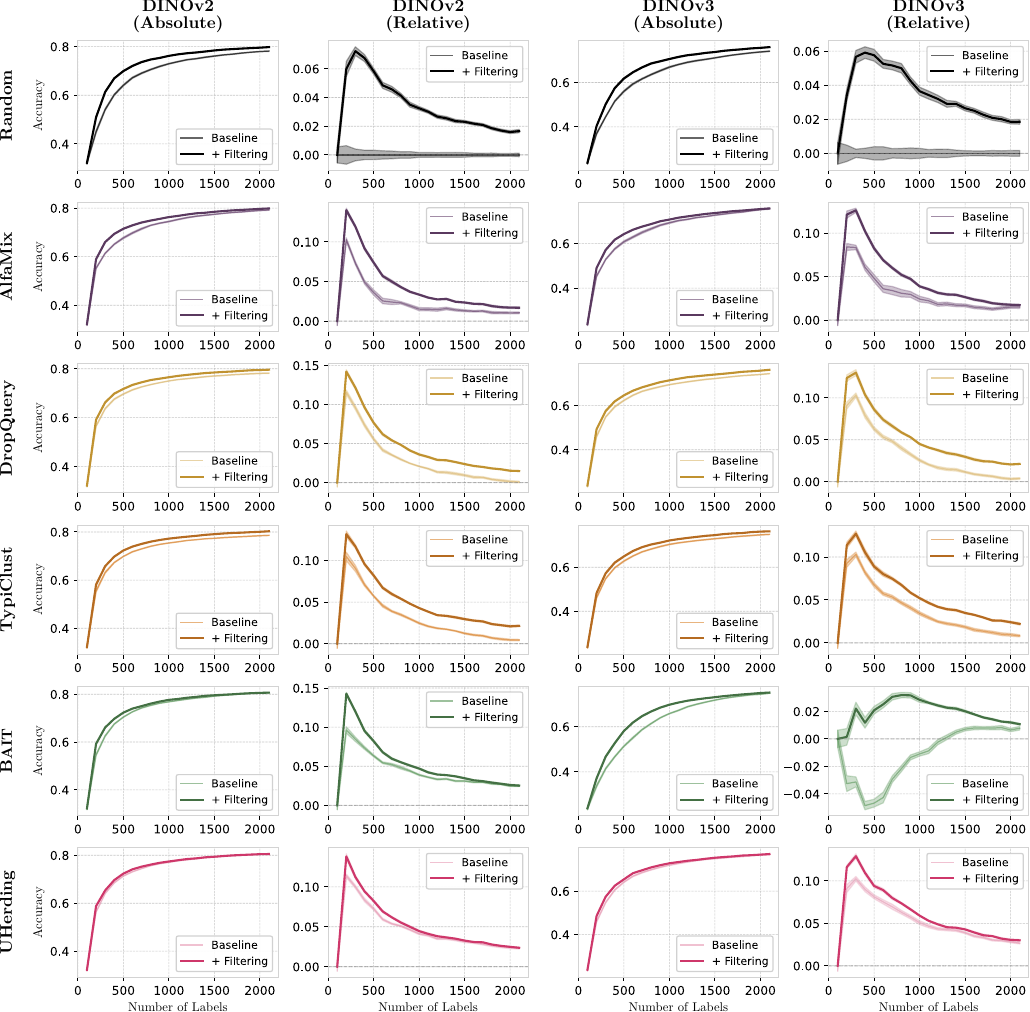}
    \caption{Relative and absolute learning curves on CIFAR-100 comparing selection strategies applied to the filtered pool (bold lines) versus the unfiltered pool (normal lines) using DINOv2 and DINOv3 backbones.}
    \label{fig:complete_strats}
\end{figure*}

\textbf{Filtering as Pool Refinement.} In \cref{fig:aulc_improvments}, we reported improvements in AULC [\%], which showed that all strategies applied to the progressively filtered pool $\mathcal{C}_R$ lead to performance improvement. For these strategies, \cref{fig:complete_strats} presents the corresponding accuracy learning curves. The figure compares strategies applied to the filtered pool versus the unfiltered pool.

\Cref{fig:complete_strats} demonstrates that progressive filtering acts as a powerful preprocessing step, consistently improving the accuracies of individual AL strategies. This behavior is consistent across both DINOv2 and DINOv3 backbones. While the impact is most substantial for poorly performing AL strategies such as AlfaMix, progressive filtering still enhances strong performers such as UHerding, even if the improvement is less pronounced. Moreover, filtering significantly improves even random sampling, demonstrating its ability to clean uninformative instances from the unlabeled pool. Furthermore, progressive filtering can act as a crucial stabilizer for failing AL strategies. For example, while standard BAIT performs worse than random sampling on DINOv3, applying filtering fixes this issue and substantially improves the strategy's effectiveness.

\begin{figure}[t]
    \centering
    \includegraphics[width=\linewidth]{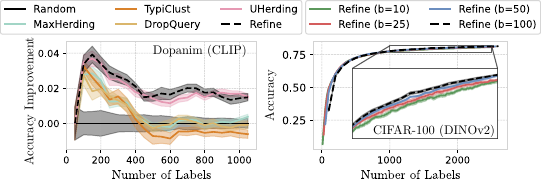}
    \caption{
    (Left) Relative learning curves of \methodname and all ensemble members on Dopanim. 
    (Right) Absolute learning curves of \methodname on CIFAR-100 for different batch sizes $b$.}
    \label{fig:robustness_and_batchsize}
\end{figure}

\textbf{Weak Ensemble Members.} We investigate a potential failure case in \methodname by considering an ensemble where the majority of strategies perform worse than random sampling. Specifically, we run \methodname on Dopanim (CLIP) using an ensemble of three weak and one strong strategy. All ensemble members and \methodname are shown in \cref{fig:robustness_and_batchsize} (left). Despite this unfavorable composition, \methodname performs well, closely following the strong ensemble member. 
A promising future direction may be to measure the similarity of ensemble selections and reduce the influence of non-complementary strategies over the AL cycles. This would encourage more independent selection within the ensemble, increasing the likelihood of retaining valuable instances.

\textbf{Varying Batch Size.} We investigate \methodname's behavior when changing the batch size $b$. Hence, we evaluate \methodname on CIFAR-100 (DINOv2) with batch sizes $b \in \{10,25,50,100\}$ under a fixed total budget of $B \approx 2500$. Interestingly, results in \cref{fig:robustness_and_batchsize} (right) show that larger batch sizes yield slightly better performance. We attribute this to the filtering process benefiting from a larger candidate set: each strategy contributes more instances, increasing the diversity of the filtered pool. Over multiple rounds, this effectively exposes the filtering process to a larger portion of the unlabeled pool, making it more likely that valuable instances are preserved.

\textbf{Alternative for second stage.}
While coverage-based selection in the second stage of \methodname performs well empirically, it is solely a single strategy applied to the refined pool $\gC_R$ without focusing on the most valuable heuristic in the current cycle. To provide an outlook on how we can realize this focus, we additionally consider a more deliberate selection process. Rather than applying a single coverage-based strategy to $\gC_R$, we use each strategy in the ensemble independently. With an ensemble of two strategies, this results in two proposed batches, each reflecting a different heuristic. The selection process then reduces to assessing which of the proposed batches is expected to yield the greatest improvement in model performance.

\begin{figure}[!ht]
    \centering
    \includegraphics[width=.9\linewidth]{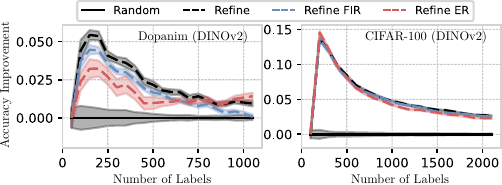}
    \caption{Relative learning curves of \methodname on Dopanim and CIFAR-100 (DINOv2) using coverage-based, FIR-based, and EER-based batch selection in the second stage.}
    \label{fig:fir_eer}
\end{figure}

To rank the candidate batches, we investigate two established criteria from the literature: the Fisher information ratio (FIR)~\cite{sourati2017asymptotic} and the expected error (ER)~\cite{roy2001toward}. Both criteria estimate the impact on model performance of including a batch in the training dataset. Notably, these criteria cannot be straightforwardly applied in a deep batch AL setting~\cite{huseljic2024fast,huseljic2025efficient}, as the search for a candidate batch is combinatorially infeasible given a large unlabeled pool. Letting each strategy in the ensemble construct a candidate batch effectively guides the search through this otherwise intractable space~\cite{huseljic2025boss}. 
The results in \cref{fig:fir_eer} demonstrate that this selection process yields comparable results to coverage-based selection, highlighting that the second stage of \methodname can easily adapted with more sophisticated components. We consider this a promising direction for future work, where such an selection mechanism could, for example, incorporate cost-sensitive criteria or realize a truly adaptive selection. For further details, we refer to our implementation.

\section{Additional Results and Experiments}\label{app:add_experiments}

\textbf{Parallel Acquisition Times.}
To provide insights into the acquisition times of \methodname and demonstrate how they can be reduced through parallelization, we report runtimes (min:sec) on CIFAR-100 (DINOv2) in \cref{tab:query_times}. Sequential acquisition times remain on the same order as training times, and at later cycles the latter actually dominates. Since the selections of individual ensemble members are independent, parallelization is straightforward.  We compute projected parallel runtimes using Amdahl's Law, assuming a conservatively estimated 85\% of the acquisition process is parallelizable, with a realistic scheduling overhead of 3\% for 8 and 8\% for 80 processes. With 8 processes, acquisition drops below training time, making it no longer the computational bottleneck. For reference, AutoAL~\cite{wang2025autoal}, another computationally intensive ensemble AL method, requires 4:24 and 8:41 at cycles 10 and 20, respectively, on the same hardware.  Unlike \methodname, AutoAL cannot be parallelized as easily, as an auxiliary model is updated after all strategies have produced their selections, limiting scalability.
\begin{table}[!ht]
    \centering
    \caption{Acquisition times (min:sec) of \methodname on CIFAR-100 (DINOv2) at AL cycles 10 and 20.}\label{tab:query_times}
    \scriptsize
    \begin{tabular}{r|l|lll}
        Cycle & Training & Seq.~Acq. & Par.~Acq.~(8$\times$) & Par.~Acq.~(80$\times$) \\
        \midrule
        10 & 07:50 & 09:58 & 02:38 & 01:44 \\
        20 & 14:34 & 10:31 & 02:47 & 01:50 \\
    \end{tabular}
\end{table}

\textbf{Aggressiveness of Progressive Filtering.}
To provide insights into how aggressively progressive filtering reduces the size of the candidate pools $\gC_r$, we report the ratio $\nicefrac{|\gC_r|}{|\gU|}$ across filtering rounds $r$ for several datasets in \cref{tab:aggresiveness}. For datasets with small batch sizes such as CIFAR-10 ($b=10$), filtering is highly aggressive: after five rounds, only approximately 5\% of unlabeled instances remain in $\mathcal{C}_R$, meaning over 95\% are discarded before final acquisition. For datasets with larger batch sizes such as CIFAR-100 and Tiny ImageNet, the reduction is notably less pronounced. This also connects to the batch size analysis in \cref{fig:robustness_and_batchsize} (right), where larger batch sizes benefit \methodname by considering a larger portion of the unlabeled pool. 
\begin{table}[!ht]
    \centering
    \scriptsize
    \caption{Ratio $\nicefrac{|\mathcal{C}_R|}{|\mathcal{U}|}$ across filtering rounds on several datasets.}\label{tab:aggresiveness}
    \begin{tabular}{r|ccccc}
    Filtering Round & 1 & 2 & 3 & 4 & 5 \\
    \midrule
    CIFAR-10 ($b=10$)      & 0.163 & 0.087 & 0.067 & 0.058 & 0.052 \\
    Dopanim ($b=50$)       & 0.453 & 0.293 & 0.232 & 0.199 & 0.179 \\
    CIFAR-100 ($b=100$)    & 0.689 & 0.415 & 0.323 & 0.277 & 0.247 \\
    Tiny ImageNet ($b=200$) & 1.000 & 0.711 & 0.586 & 0.516 & 0.472 \\
    \end{tabular}
\end{table}

\begin{figure}[!t]
    \centering
    \includegraphics[width=\linewidth]{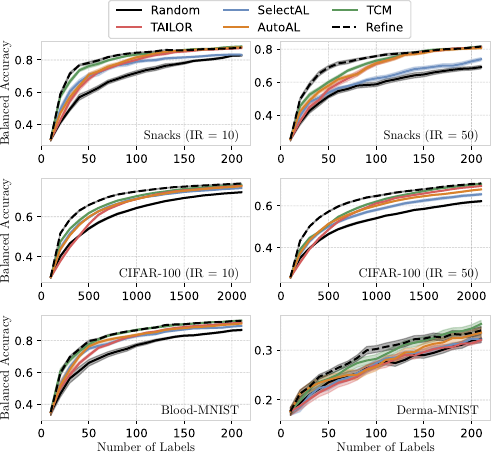}
    \caption{Absolute learning curves of ensemble AL methods on long-tailed and out-of-distribution datasets using DINOv2.}\label{fig:lt_ood_lcs}
\end{figure}

\textbf{Robustness to Class Imbalance \& Distribution Shift.} To further investigate the robustness of \methodname in more realistic settings, we conduct additional experiments considering both class imbalance and out-of-distribution data, \ie, data that strongly differs from the foundation model's training distributions. For the class-imbalance setting, we run \methodname (DINOv2) on long-tailed versions of Snacks and CIFAR-100 using exponential class decay with imbalance ratios $\text{IR} \in \{10, 50\}$, denoting the ratio of the largest to the smallest class. For the out-of-distribution setting, we include BloodMNIST and DermaMNIST from the MedMNIST benchmark~\cite{medmnistv2}. The results in \cref{fig:lt_ood_lcs} show that \methodname maintains strong performance, demonstrating robustness across both  settings.

\begin{figure}
    \centering
    \includegraphics[width=\linewidth]{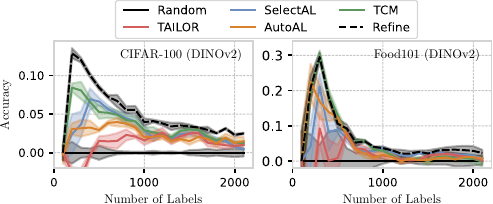}
    \caption{Relative learning curves of ensemble AL methods when fine-tuning the last two transformer blocks of DINOv2.}
    \label{fig:finetune}
\end{figure}

\textbf{Fine-tuning Experiments.}
While the main experiments employ frozen features, reflecting current practice~\cite{hacohen2022active,huseljic2024fast,huseljic2025efficient}, \methodname is agnostic to the downstream model. To verify this, we investigate the performance of ensemble AL methods when fine-tuning the last two transformer blocks of a DINOv2 pretrained transformer. This setting is more challenging, as training hyperparameters, \ie, learning rate and weight decay, had to be tuned on CIFAR-10 and transferred to the other datasets. Since the labeled pool grows in each AL cycle, these hyperparameters are highly likely to become suboptimal over the course of the AL process~\cite{munjal2022towards,huseljic2023role}. The relative learning curves in \cref{fig:finetune} show that \methodname maintains strong performance.

\section{Main Results: All Learning Curves}\label{app:complete_lcs}
\begin{figure*}
    \centering
    \includegraphics[width=.75\linewidth]{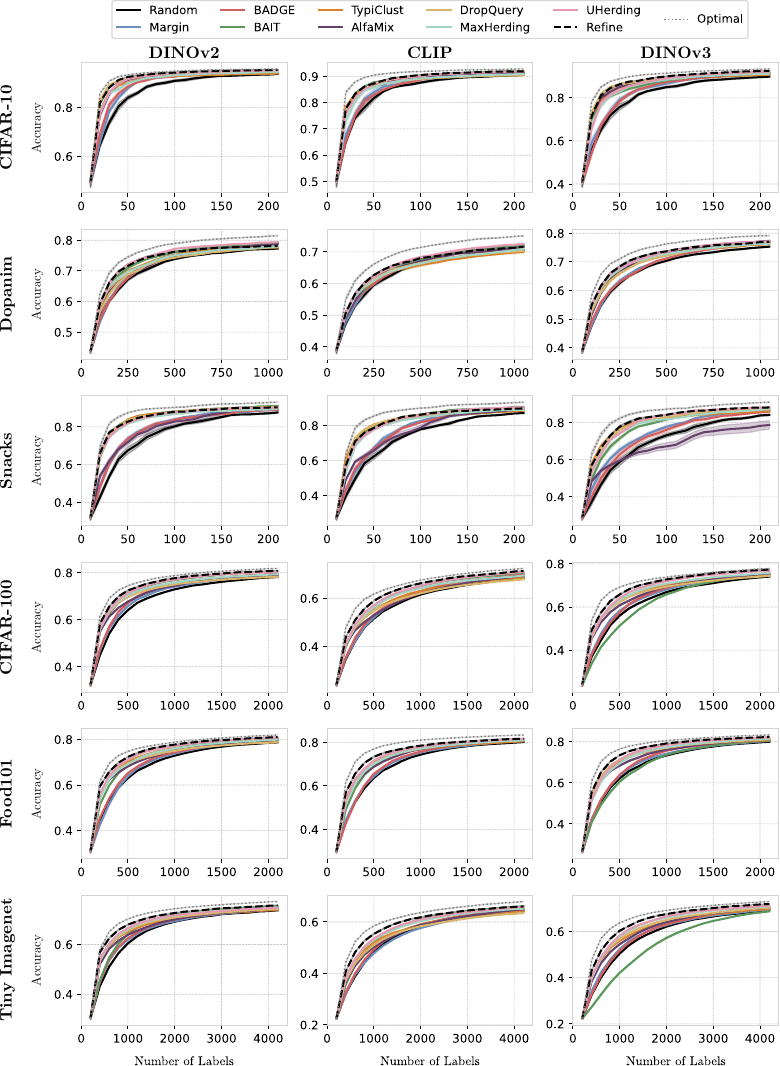}
    \caption{Complete set of absolute learning curves for individual AL strategies showing the accuracy for all datasets and backbone combinations. The optimal performance is approximated using BoSS~\cite{huseljic2025boss}.}
    \label{fig:complete_absolute}
\end{figure*}

\begin{figure*}
    \centering
    \includegraphics[width=.75\linewidth]{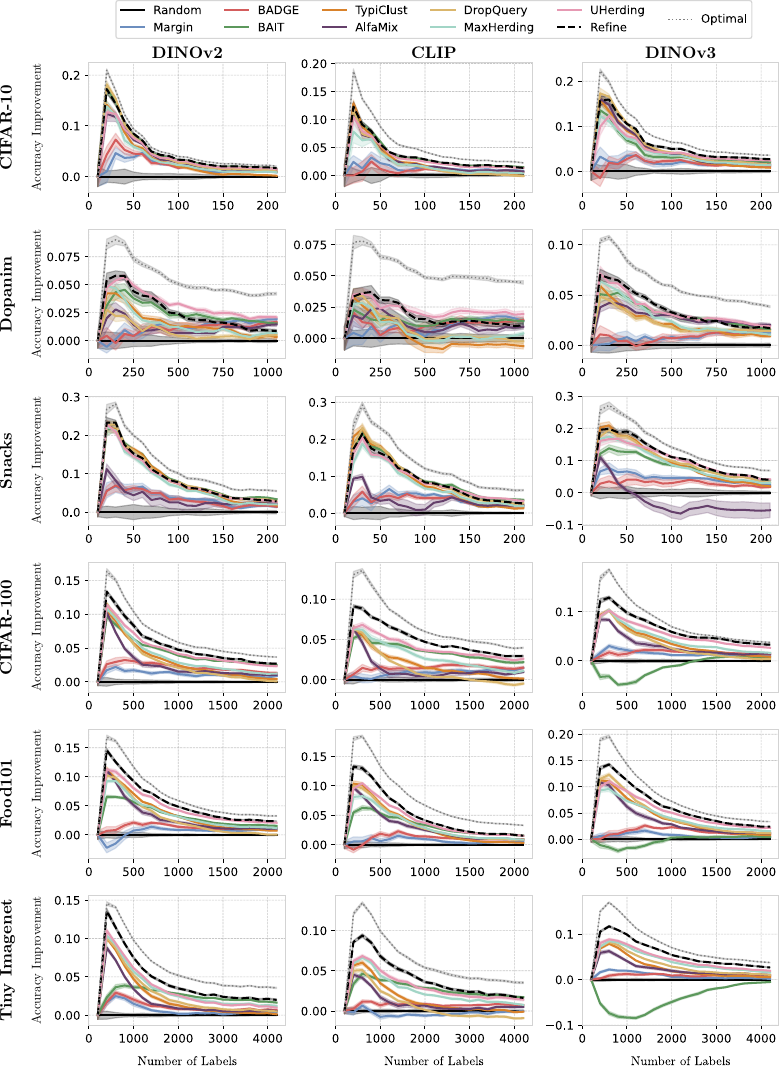}
    \caption{Complete set of relative learning curves for individual AL strategies showing the accuracy relative to random sampling from $\gU_t$ for all datasets and backbone combinations. The optimal performance is approximated using BoSS~\cite{huseljic2025boss}.}
\label{fig:complete_relative}
\end{figure*}

\begin{figure*}
    \centering
    \includegraphics[width=.75\linewidth]{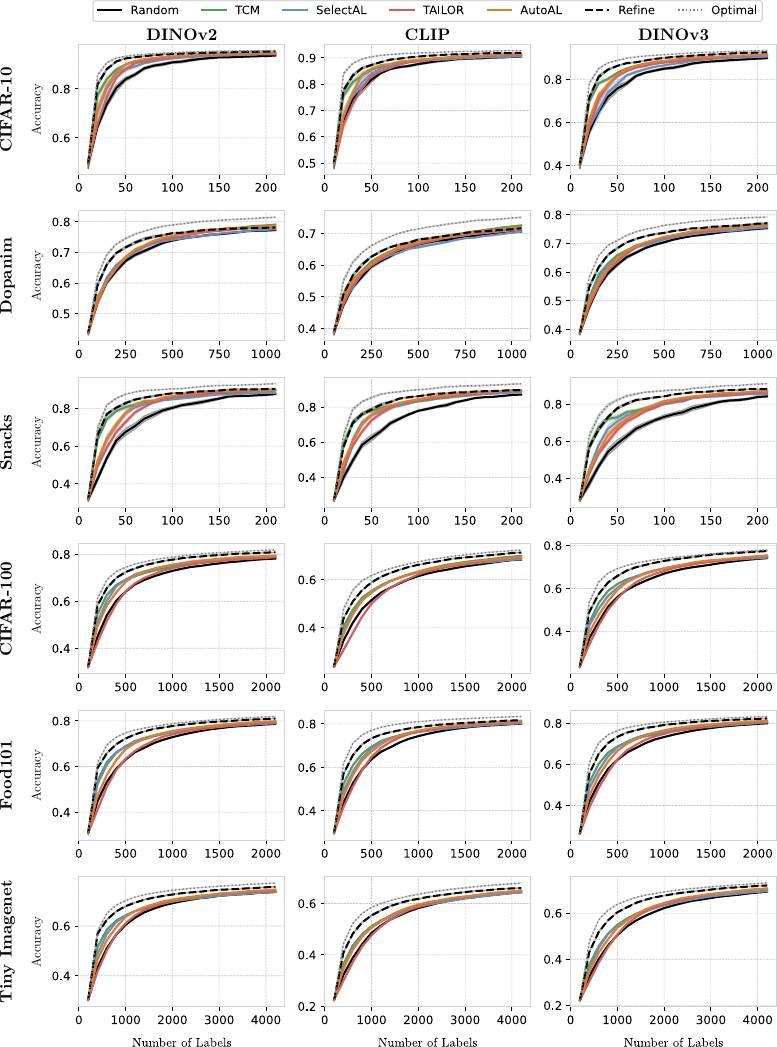}
    \caption{Complete set of absolute learning curves for ensemble AL methods showing the accuracy for all datasets and backbone combinations. The optimal performance is approximated using BoSS~\cite{huseljic2025boss}.}
    \label{fig:complete_abs_ensemble}
\end{figure*}

\begin{figure*}
    \centering
    \includegraphics[width=.75\linewidth]{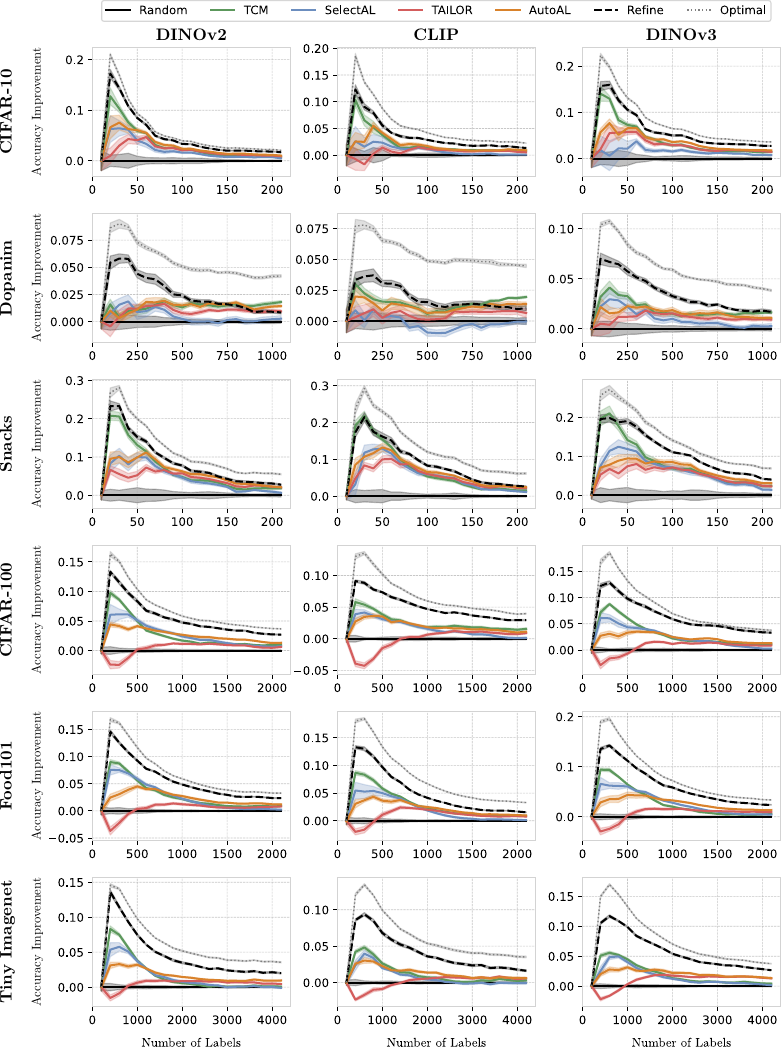}
    \caption{Complete set of relative learning curves for ensemble AL methods showing the accuracy relative to random sampling from $\gU_t$ for all datasets and backbone combinations. The optimal performance is approximated using BoSS~\cite{huseljic2025boss}.}
\label{fig:complete_rel_ensemble}
\end{figure*}

This section presents the complete set of learning curves used to derive \cref{fig:pairwise,fig:pairwise_ensemble}, which were omitted from the main paper due to space limitations. For individual AL strategies, \cref{fig:complete_absolute} displays the absolute learning curves (accuracy), while \cref{fig:complete_relative} shows the relative learning curves (accuracy relative to random sampling from $\gU_t$). Similarly, for ensemble AL methods, \cref{fig:complete_abs_ensemble,fig:complete_rel_ensemble} present the absolute and relative learning curves, respectively. These results provide a comprehensive overview supporting the conclusions drawn in the main text. 
Additionally, we include BoSS~\cite{huseljic2025boss}, an oracle strategy that approximates the optimal selection strategy by leveraging label information unavailable to standard AL strategies. By comparing against this oracle, we can assess how closely the AL strategies and \methodname approach the theoretical optimum.

\Cref{fig:complete_absolute,fig:complete_relative} demonstrate that \methodname outperforms all other individual AL strategies across nearly every dataset and backbone, including DINOv2, CLIP, and DINOv3. It excels in both low- and high-budget scenarios, consistently delivering high accuracies throughout the entire AL process. Furthermore, \methodname proves to be robust, maintaining a strong lead across diverse datasets while remaining invariant to the choice of the backbone. Similarly, \Cref{fig:complete_abs_ensemble,fig:complete_rel_ensemble} demonstrate that \methodname outperforms all ensemble AL methods. These approaches generally struggle to effectively combine multiple strategies, often failing to exceed the performance of individual AL strategies. With \methodname, we propose the first ensemble AL method capable of effectively combining multiple AL strategies into a unified framework that consistently surpasses individual baselines. This improvement over existing ensemble AL methods is statistically significant ($\alpha = 0.01$), as confirmed by a Friedman test~\cite{friedman1937use} on the AULC across 18 blocks (dataset $\times$ model) followed by post-hoc paired Wilcoxon signed-rank tests~\cite{wilcoxon1945individual} with Holm correction~\cite{holm1979simple}.

When considering the optimal strategy approximated by the BoSS oracle, we observe that most strategies exhibit a considerable gap to this optimum. Notably, \methodname closes this gap more effectively than other strategies, yet a significant gap to the oracle remains, underscoring the need for continued research into more effective AL selection strategies.


    
\end{document}